\PassOptionsToPackage{table,dvipsnames}{xcolor}
\documentclass{article}
\usepackage{iclr2027_conference,times}
\usepackage[T1]{fontenc}
\usepackage{fancyhdr}
\usepackage{amsmath,amsfonts,bm}

\def\eqref#1{equation~\ref{#1}}
\def\1{\bm{1}}

\DeclareMathAlphabet{\mathsfit}{\encodingdefault}{\sfdefault}{m}{sl}
\SetMathAlphabet{\mathsfit}{bold}{\encodingdefault}{\sfdefault}{bx}{n}

\usepackage{microtype,mathtools,amssymb,bm,xspace}
\usepackage{graphicx,booktabs,array,tabularx,multirow}
\usepackage{tikz}
\usepackage{eso-pic}
\usetikzlibrary{arrows.meta,positioning,calc}
\usepackage{placeins,flafter,wrapfig,capt-of}
\makeatletter
\newcommand{\finishwrap}{\par
  \ifnum\c@WF@wrappedlines>1\relax
    \vspace{\dimexpr\c@WF@wrappedlines\baselineskip-2\baselineskip\relax}
  \fi
  \WFclear}
\makeatother
\definecolor{MethodBlue}{HTML}{244F72}
\definecolor{MethodGreen}{HTML}{286458}
\definecolor{SoftBlue}{HTML}{EFF4F8}
\definecolor{SoftGreen}{HTML}{EFF6F3}
\definecolor{SoftGray}{HTML}{F3F4F5}

\newcolumntype{Y}{>{\centering\arraybackslash}X}
\newcolumntype{L}{>{\raggedright\arraybackslash}X}

\usepackage{hyperref}
\hypersetup{colorlinks=true,linkcolor=MethodBlue,citecolor=MethodGreen,urlcolor=MethodBlue,
pdftitle={Positive-Unlabeled Learning for Agent Safety False Alarm Auditing},
pdfauthor={Xichen Yan, Chongyang Gao, Kezhen Chen, Guangyi Zhang, Jiaqi Wu, Lixu Wang}}
\usepackage[capitalize,noabbrev]{cleveref}

\makeatletter
\def\label@in@display#1{%
  \ifx\df@label\@empty\gdef\df@label{#1}\fi
}
\newcommand{\relaylabelalias}[2]{%
  \@ifundefined{r@#2}{}{%
    \expandafter\let\csname r@#1\expandafter\endcsname\csname r@#2\endcsname
    \expandafter\let\csname r@#1@cref\expandafter\endcsname\csname r@#2@cref\endcsname
  }%
}
\AddToHook{begindocument/end}{%
  \relaylabelalias{eq:complete-reference}{eq:method:tn-anchor}
  \relaylabelalias{eq:complete-protected-loss}{eq:method:protected-comparison}
  \relaylabelalias{eq:complete-reference-ranks}{eq:method:rank-consensus}
  \relaylabelalias{eq:complete-distillation}{eq:method:rank-distillation}
  \relaylabelalias{eq:complete-training}{eq:method:student-objective}
  \relaylabelalias{eq:complete-hierarchy}{eq:method:hierarchical-increment}
  \relaylabelalias{eq:complete-final}{eq:method:final-score}
}
\makeatother

\title{Positive-Unlabeled Learning for Agent Safety False Alarm Auditing}
\newcommand{\authoremail}[1]{\href{mailto:#1}{\textcolor{black}{\texttt{#1}}}}
\author{%
\begin{tabular}[t]{@{}l@{}}
\textbf{Xichen Yan}\textsuperscript{1,*},
\textbf{Chongyang Gao}\textsuperscript{2,3},
\textbf{Kezhen Chen}\textsuperscript{3},
\textbf{Guangyi Zhang}\textsuperscript{4},
\textbf{Jiaqi Wu}\textsuperscript{5},
\textbf{Lixu Wang}\textsuperscript{6,\textdagger}\\[6pt]
{\normalfont\normalsize\textsuperscript{1}Jinan University,
\textsuperscript{2}Northwestern University,
\textsuperscript{3}Analogy AI, Inc.},
{\normalfont\normalsize\textsuperscript{4}Shenzhen Technology University}\\
{\normalfont\normalsize\textsuperscript{5}Tsinghua University},
{\normalfont\normalsize\textsuperscript{6}The Chinese University of Hong Kong, Shenzhen}\\[7pt]
{\normalfont\small\authoremail{xxy315@student.bham.ac.uk}\quad
\authoremail{chongyanggao2026@u.northwestern.edu}}\\[1pt]
{\normalfont\small\authoremail{kzchen0204@gmail.com}\quad
\authoremail{zhangguangyi@sztu.edu.cn}}\\[1pt]
{\normalfont\small\authoremail{wujiaqi@mail.tsinghua.edu.cn}\quad
\authoremail{wanglixu@cuhk.edu.cn}}
\end{tabular}%
}
\iclrfinalcopy
\fancypagestyle{preprintfirst}{%
  \fancyhf{}
  \fancyfoot[C]{\thepage}
  \renewcommand{\headrulewidth}{0pt}
}
\makeatletter
\def\@maketitle{%
  \vbox{\hsize\textwidth
    {\LARGE\scshape\@title\par}
    \vspace{0.22in}
    {\normalfont\normalsize\@author\par}
    \vspace{0.30in}
  }%
}
\makeatother

\renewcommand{\headrulewidth}{0.4pt}

\begin{document}
\maketitle
% \fancyhf{}
% % \fancypagestyle{preprintpages}{
% % \pagestyle{fancy}
% \fancyhead[L]{\itshape Preprint. Under review.}
% \fancyfoot[C]{\thepage}
% \renewcommand{\headrulewidth}{0.4pt}

\thispagestyle{preprintfirst}
\AddToShipoutPictureFG*{%
  \begin{tikzpicture}[remember picture,overlay]
    \node[anchor=north west,inner sep=0pt]
      at ([xshift=1.5in,yshift=-0.02in]current page.north west)
      {\includegraphics[width=1.8in]{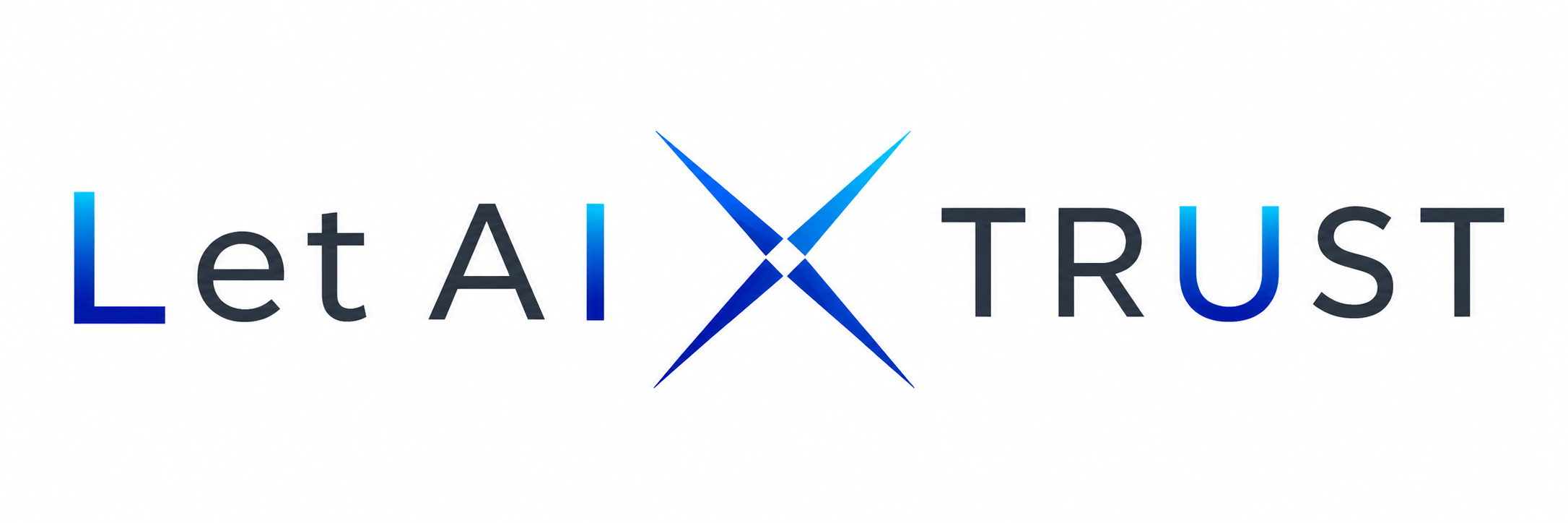}};
    \draw[line width=0.4pt]
      ([xshift=1.5in,yshift=-0.55in]current page.north west) --
      ([xshift=7in,yshift=-0.55in]current page.north west);
  \end{tikzpicture}%
}
\begingroup
\renewcommand{\thefootnote}{\fnsymbol{footnote}}
\footnotetext[1]{Work carried out during a research internship at The Chinese University of Hong Kong, Shenzhen.}
\footnotetext[2]{Corresponding author.}
\endgroup
\raggedbottom
\begin{abstract}
Safety monitors help safeguard language-model agents interacting with external tools and environments, but conservative monitoring can generate many false alarms, consuming extensive review resources and weakening trust in alerts.
Because false and genuine alarms often remain interleaved in native monitor scores, obtaining a reliable cutoff still requires substantial manual verification.
In practice, a small set of verified-safe non-alarmed trajectories may be available while alarms remain unlabeled, naturally casting false-alarm auditing as a positive--unlabeled (PU) ranking problem.
The key challenge is \emph{monitor-induced selection}, since observed safe references are accepted by the monitor, while the hidden safe alarms of interest are precisely those it incorrectly flags, making the observed positives poorly representative of the positives to be recovered.
To address this challenge, we propose a two-stage framework in which \emph{Trust-aware PU Supervision} adapts safe references toward the alarm domain and protects plausible false alarms from excessive negative pressure, while \emph{Reliability-gated Rank Distillation} consolidates consistent ordering preferences from multiple PU reference models into a single student.
\emph{Consensus-guided Structural Refinement} then improves the student ranking using hierarchical safe-reference support, alarm relations, and predicted reference consensus.
The framework requires no alarm safety labels for fitting and leaves the underlying monitor unchanged.
Across mainstream safety monitors, our method achieves a macro AUPRC of $0.6444$, outperforming eight evaluated PU baselines by $5.27$--$16.98$ absolute percentage points; compared with PULDA, the strongest evaluated PU baseline, it recovers $33.3\%$ more false alarms at a $5\%$ review budget.
Code is available at \url{https://anonymous.4open.science/r/safe-alarm-ranking-c4d2/}.
\end{abstract}
\section{Introduction}
\label{sec:introduction}

\begin{figure}[!t]
\centering
\includegraphics[width=\linewidth]{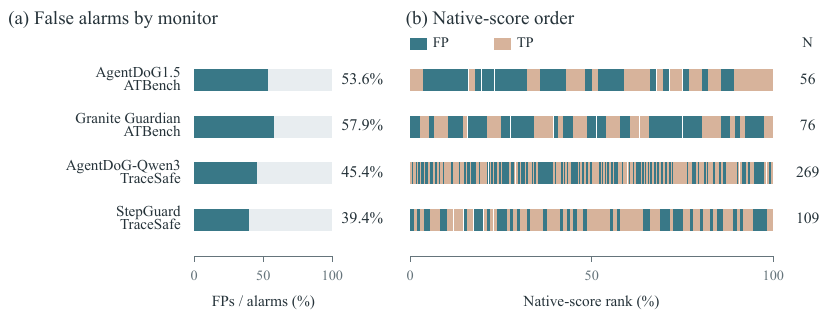}
\caption{\textbf{False alarms and overlap in the native score order.} (a) Share of safe trajectories among alarms in four dataset--monitor settings.
(b) The native monitor-score ordering for the same four settings, with lower-risk alarms on the left.
Each tile is one alarm (teal for FP, sand for TP); row widths are normalized to cohort size, and $N$ gives the alarm count.
All $510$ alarms are shown.
FP and TP remain interleaved within each order, leaving no clean score cutoff.
Ground-truth labels determine tile colors only, not the ordering; ties in native scores are broken by sample ID.}
\label{fig:intro-diagnostics}
\end{figure}

Language-model agents increasingly perform real-world tasks through multi-round interactions with external tools and environments.
During execution, they may be exposed to prompt injection attacks or take actions that violate safety constraints, potentially causing serious harm \citep{debenedetti2024agentdojo,wang2021non,sun2026survey,jiao2025can,feng2025token}.
An agent's execution trajectory records its intermediate actions, observations, and tool interactions, providing a natural basis for safety monitoring mechanisms that detect unsafe behavior \citep{liu2026agentdog,xiang2025guardagent,chen2025shieldagent}.
Existing trajectory monitors assess such execution traces using learned evaluators or explicit safety criteria and produce safety scores, risk judgments, or intervention signals when potentially unsafe behavior is detected.
These approaches have demonstrated effective detection of unsafe agent behavior across a range of settings, including coding, web interaction, and tool-use tasks \citep{liu2026agentdog,xiang2025guardagent,chen2025shieldagent,li2026webcloak}.

%第二段改前：
%\lixu{However, such high recall is problematic.} Safety monitors are often designed conservatively to avoid overlooking 
%potentially unsafe behavior. Such conservatism can arise not only from
%sensitive alarm thresholds, but also from broad safety policies and
%uncertainty in distinguishing similar actions under benign and unsafe
%contexts. As a result, benign trajectories may also trigger alarms,
%increasing the proportion of false positives (FPs) in the alarm pool\lixu{directly state the problem}.
%Such failures have been observed in agent-defense evaluations: AgentDojo
%reports that false positives from an evaluated prompt-injection detector
%degrade benign utility \citep{debenedetti2024agentdojo}. \lixu{Our findings...}
%Frequent false alarms can undermine users' trust in warnings and reduce
%their willingness to respond, potentially weakening responses to genuine
%alarms \citep{chancey2017trust}. They also consume human and computational
%resources for investigation, verification, and response; under a fixed
%review and response budget, resources spent on false alarms leave less
%capacity for genuine safety violations
%\citep{alahmadi2022falsepositives,oliver2024carbonfilter}.
%To examine whether this problem is substantial in agent safety monitoring,
%we analyze four dataset--monitor settings and find that
%$39.4\%$--$57.9\%$ of raised alarms correspond to safe trajectories
%(\cref{fig:intro-diagnostics}a).

%第二段改后：

However, this seemingly strong performance in unsafe-behavior detection comes with a substantial cost.
Agent trajectories vary considerably in structure and format across tasks and datasets, while the safety of an individual action often depends on both the task context and the preceding execution history \citep{yuan2024rjudge,chen2026tracesafe}.
To reduce the risk of overlooking unsafe behavior across such heterogeneous settings, safety monitors often favor sensitivity, for example by adopting permissive alarm thresholds or overly inclusive safety criteria that also flag borderline cases.
Our empirical study reveals that this design choice results in a substantial number of false alarms: across mainstream safety monitors, $39.4\%$--$57.9\%$ of triggered alarms actually correspond to safe trajectories and should therefore be regarded as false positives (FPs) (\cref{fig:intro-diagnostics}a).
Such a high false-positive rate is consequential.
Repeated false warnings can erode users' trust in the monitoring system and reduce their willingness to respond to subsequent alerts \citep{chancey2017trust}.
Moreover, each false alarm must be investigated before it can be dismissed, incurring additional human effort and computational overhead \citep{alahmadi2022falsepositives,yang2024trueattacks,oliver2024carbonfilter}.

%修改前的第三段：
%A direct way to mitigate the burden\lixu{the burden???} of false alarms is to reuse the safety
%monitor's native confidence scores\lixu{what is the confidence, you don't introduce}, either to prioritize low-confidence
%alarms for review or to suppress them with an additional threshold.
%Native-score review is useful only if FPs are sufficiently concentrated
%near the front of the queue; suppressing alarms additionally requires a
%cutoff whose effect on genuine violations can be validated. However, the
%native score measures how strongly a trajectory appears unsafe, whereas
%the auditing problem asks a different conditional question: given that an
%alarm has already been raised, how likely is the alarm to be wrong?
%Because both FPs and TPs have already crossed the monitor's alarm
%boundary\lixu{Too tedious!}, their native scores can substantially overlap, and the original
%ranking need not prioritize FPs. For one monitored setting, the native
%order surfaces 24 FPs among the first 54 alarms, while our method surfaces
%31\lixu{why can you present your method? you haven't introduced your methods!} (\cref{fig:intro-audit-queue}). More importantly, even when a useful
%cutoff exists for a particular alarm pool, identifying and validating it
%requires labeled alarms to determine how many false alarms are removed and
%how many genuine violations are inadvertently suppressed\lixu{the human effort cost!}.

%修改后的第三段：

A natural way to prioritize likely false alarms for review is to reuse the monitor's original safety score, possibly with an additional cutoff.
Our empirical study shows that false and true alarms often receive similar scores and remain mixed in the native ordering (\cref{fig:intro-diagnostics}b).
With this overlap, moving the cutoff cannot reliably separate the two groups.
Even if stronger future monitors produced better-separated scores, choosing a cutoff that preserves genuine alarms would require manually verifying many cases.
That verification is impractical under a limited review budget.

Let us reconsider the alarm-isolation problem from a different perspective.
In practice, it is often easy to obtain a small set of verified-safe trajectories, either from routine benign traces that do not trigger alarms or from the monitor's training data.
Because safety monitors are typically conservative, trajectories that they confidently classify as safe provide reliable positive examples.
Treating safety as the class of interest, these verified-safe trajectories form the observed positive set, while false alarms hidden within the alarm pool constitute latent positives.
The task therefore becomes one of identifying positive-class samples from an otherwise unlabeled alarm pool, naturally giving rise to a positive-unlabeled (PU) learning formulation \citep{elkan2008positive}.
However, this setting differs fundamentally from standard PU learning.
The safety monitor itself has already partitioned the two types of safe trajectories across its alarm boundary: observed positives correspond to safe behavior that the monitor accepts, whereas the latent positives of interest are precisely the safe trajectories that it incorrectly rejects.
Consequently, the observed positive set may not be representative of the false alarms in the unlabeled pool.
Learning directly from these positives without accounting for this selection mechanism risks reproducing the monitor's existing decision pattern rather than correcting its errors \citep{bekker2018sar}.

In this work, to prioritize false-alarm agent trajectories identified by safety monitors, we propose a two-stage positive--unlabeled ranking framework that can prioritize most-likely false alarms in an auditing queue.
In the first stage, we adapt the verified-safe non-alarms towards the alarm pool and reduce excessive negative pressure on plausible false alarms to mitigate the shift between the safe references and the false alarms, which helps us develop much more effective PU classifiers.
But due to the uncertainty of PU learning process, we found that trained classifiers with different initializations provide distinctive auditing queues.
Therefore, we propose a reliability-gated distillation strategy to distill the consistent alarm orderings across multiple PU classifiers into a single student.
Then, in the second stage, we can obtain an initial auditing queue by ranking the scores predicted by the student PU classifier.
To further exploit structure across task contexts, we refine the auditing queue with consensus-guided structural refinement driven by hierarchical safe-reference support, target-alarm relations, and reference agreement.
Extensive experiments on two benchmark datasets and four dataset--monitor settings demonstrated the superiority of our proposed framework in prioritizing false alarms of agent trajectories compared with a variety of state-of-the-art potential baseline approaches.
In summary, our main contributions are:

{
\setlength{\leftmargini}{1.25em}
\setlength{\labelsep}{0.4em}

\begin{itemize}
\setlength{\itemsep}{0.25em}
\setlength{\topsep}{0.25em}
\setlength{\parsep}{0pt}
\setlength{\parskip}{0pt}

\item We identify the issue of frequent false alarms raised by current agent safety monitors, and to our best knowledge, propose the first effective framework to prioritize most-likely false alarms in the auditing queue.

\item We formulate agent false-alarm auditing as a special positive--unlabeled ranking problem under monitor-conditioned selection and develop a series of designs to mitigate the shift between positives and hidden positives and search for consistency across PU learning processes and task contexts.

\item We conduct extensive experiments against established PU methods, direct auditing controls based on native monitor scores and frozen representations, and local LLM re-judgment.
Across mainstream safety monitors, our method achieves a macro AUPRC of $0.6444$, outperforming eight evaluated PU baselines by $5.27$--$16.98$ absolute percentage points and recovering $33.3\%$ more false alarms than PULDA at only a $5\%$ auditing budget (Table~\ref{tab:main-results}).

\end{itemize}
}
\section{Preliminaries}
\label{sec:related-work}
\label{sec:preliminaries}

\paragraph{Safety monitoring and LLM-based judgment.}
Existing work assesses agent safety through learned risk prediction, policy verification, and direct LLM judgment.
Granite Guardian detects risks in model inputs and outputs \citep{padhi2025graniteguardian}, while AgentDoG provides contextual risk diagnosis for execution trajectories \citep{liu2026agentdog}.
StepGuard checks candidate tool actions before execution and balances safe and unsafe supervision to reduce over-defense and under-defense \citep{zheng2026stepguard}.
Policy-based approaches include GuardAgent, which translates safety requirements into executable checks \citep{xiang2025guardagent}, and ShieldAgent, which combines policy modeling with logical verification \citep{chen2025shieldagent}.
R-Judge evaluates the ability of general-purpose LLMs to judge safety from interaction records \citep{yuan2024rjudge}.
These studies focus on improving safety judgments and interventions.
Our work addresses a complementary task after alarms are raised: using limited verified-safe supervision to prioritize likely false alarms for human review.

\paragraph{Positive--unlabeled learning.}
PU methods learn from labeled positives and an unlabeled mixture through risk estimation, distribution alignment, or sample-level supervision.
Prior work includes nnPU, which uses non-negative risk estimation to reduce overfitting \citep{kiryo2017nnpu}, and DC-PU, which constrains positive and negative risks to balance classification errors \citep{li2025dcpu}.
Dist-PU aligns predicted label distributions with class-prior information \citep{zhao2022distpu}, and PULDA combines this alignment with functional-margin learning \citep{jiang2023pulda}.
Robust-PU iteratively selects reliable negative examples \citep{zhu2023robustpu}, PUL-CPBF constructs supervision using classifiers with different probability boundaries \citep{li2024pulcpbf}, and HolisticPU identifies unlabeled examples through their prediction trends \citep{wang2023holisticpu}.
NcPU addresses unreliable PU supervision through representation alignment~\citep{zhao2026ncpu}.
Related work also addresses nonrandom positive selection \citep{bekker2018sar} and AUC optimization from PU data \citep{sakai2018puauc}.
Our setting connects these directions through monitor-induced selection: verified-safe non-alarms and hidden safe alarms share a safety label but are selected by opposite monitor decisions.
Our method adapts safe supervision and ranking refinement to recover false alarms under a fixed review budget.

\paragraph{Setting and observed data.}
Let $\tau_i$ denote an agent execution trajectory and
$y_i\in\{+1,-1\}$ its safety label, with $+1$ indicating safe
behavior and $-1$ indicating unsafe behavior.
A safety monitor $M_\psi$ with fixed configuration $\psi$
provides a representation $\mathbf h_i\in\mathbb R^d$ and
an alarm decision $a_i\in\{0,1\}$, where $a_i=1$ indicates
an alarm.
We observe a small set $\mathcal D_P$ of verified-safe
non-alarmed trajectories ($y_i=+1$, $a_i=0$), a set
$\mathcal D_U$ of unlabeled training alarms, and a separate
target cohort $\mathcal C=\{\tau_i\}_{i=1}^{N}$ with
$a_i=1$ for every member of both alarm sets.
Safety is the positive class in our PU formulation;
members of $\mathcal D_P$ are observed positives and are
termed \emph{TN references} relative to the monitor's
alarm decision.
Frozen representations, task features, and isolation-group
identifiers are available for all three sets.
Safety labels are observed only for $\mathcal D_P$ during
fitting; the labels and false-alarm proportions of
$\mathcal D_U$ and $\mathcal C$ are unknown.

\paragraph{Problem definition.}
The auditor produces an ordering of the $N$ target alarms
for human review.
Under a review budget $B\in\{1,\ldots,N\}$, only the first
$B$ alarms are inspected, and the objective is to maximize
the number of false alarms ($y_i=+1$) among them.
The ordering may depend on the unlabeled target features;
target safety labels are used only for evaluation after
the order is fixed.
The auditor determines review priority while preserving
all original alarm decisions.

\section{Method}
\label{sec:method}
\subsection{Overview}
\label{sec:method:overview}
\begin{figure}[!htbp]
\centering
\includegraphics[width=\linewidth]{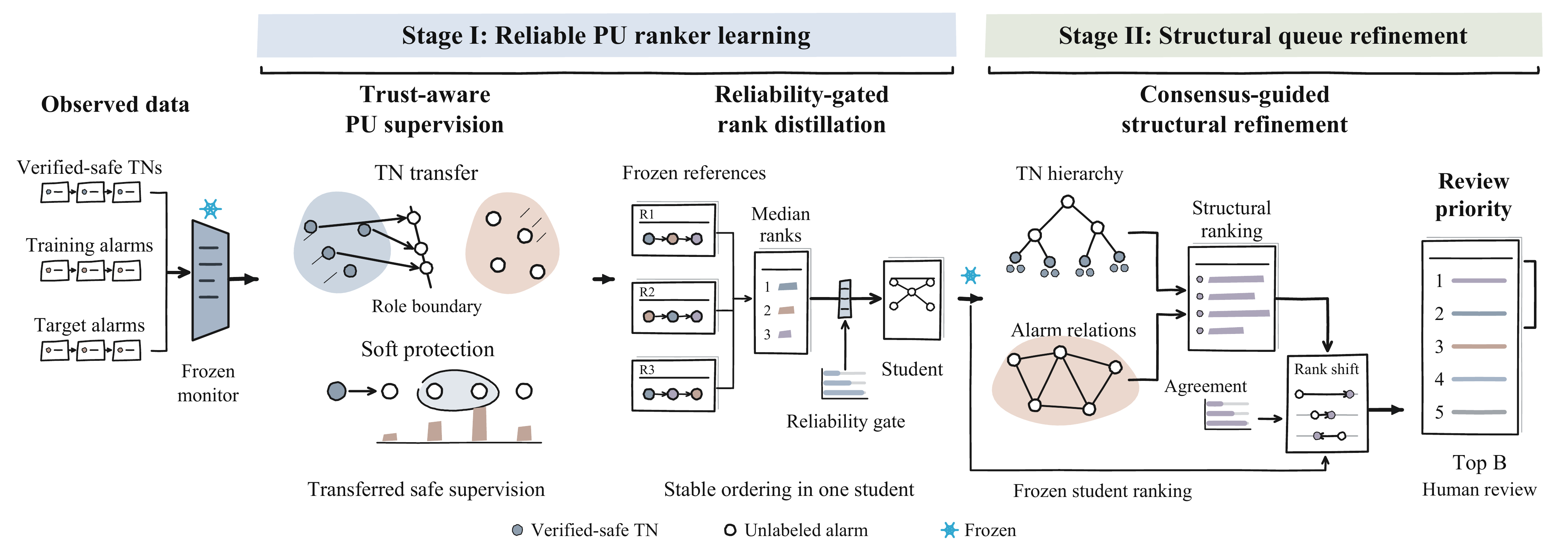}
\caption{\textbf{Overview of our method.} Stage~I learns one student through trust-aware TN transfer and soft protection (\S\ref{sec:method:trust-aware-pu}), followed by median-rank supervision with a shared reliability gate (\S\ref{sec:method:rank-distillation}).
Stage~II freezes the student, combines hierarchical TN support and target-alarm relations into a structural ranking, and uses an agreement head to scale the final rank shift (\S\ref{sec:method:stage-two}).
Components 1--2 form Stage~I and component 3 forms Stage~II; the final queue uses no test-time reference ensemble.}
\label{fig:relaypu-overview}
\end{figure}

The difficulty is that the known-safe trajectories passed the monitor, whereas the false alarms we seek were selected by it.
Treating every alarm as a negative example can therefore bury the very trajectories that reviewers should see first.
Stage~I learns a PU auditor in two steps.
Trust-aware supervision transfers TN references toward the alarm domain and reduces negative pressure on plausible FPs.
Reliability-gated distillation then combines the ordering preferences of several frozen reference auditors in one student.
Stage~II freezes that student and refines its queue with hierarchical TN support and relations among target alarms.
A predicted measure of reference agreement controls the final per-alarm adjustment.
Both stages may use unlabeled target features; neither uses target safety labels for fitting.
Figure~\ref{fig:relaypu-overview} follows these three components from left to right.

\subsection{Stage I: Reliable PU Ranker Learning}
\label{sec:method:stage-one}
Precursor, reference, and student auditors share a PU objective.
Its common part regularizes predicted safety scores across verified-safe and alarm batches through a distribution-alignment loss $\mathcal L_{\mathrm{align}}$, and adds a group-structured representation mixing loss $\mathcal L_{\mathrm{mix}}$: $\mathcal L_{\mathrm{base}} =\mathcal L_{\mathrm{align}} +\lambda_{\mathrm{mix}}(t)\mathcal L_{\mathrm{mix}}$.
Here $t$ is the training step and $\lambda_{\mathrm{mix}}(t)$ is its mixing weight.
Appendix~\ref{app:complete-base} gives the terms and optimization.
The two components below determine how safe references and other alarms contribute to fitting.

\subsubsection{Trust-Aware PU Supervision}
\label{sec:method:trust-aware-pu}
\paragraph{Transferring safe references.} A role classifier, fitted out of fold on training data, distinguishes TN references from training alarms using monitor and task representations.
Its decision boundary describes the monitor's \emph{selection} of trajectories into the alarm pool; it is not a boundary between safe and unsafe behavior.
For a TN reference $\tau_p\in\mathcal D_P$, let $\widetilde{\mathbf h}_p$ be its nearest feasible representation on that role boundary, with the task representation and monitor-vector norm held fixed.
A transfer weight $\lambda_p^{\mathrm{tr}}\in[0,1]$ reflects boundary proximity and the displacement needed to reach it.
The positive comparison score is
\begin{equation}
s_p^{\mathrm{ref}}
=(1-\lambda_p^{\mathrm{tr}})f_\theta(\mathbf h_p)
 +\lambda_p^{\mathrm{tr}}f_\theta(\widetilde{\mathbf h}_p).
\label{eq:method:tn-anchor}
\end{equation}
Thus a well-supported transfer uses more of the boundary-oriented score, while a doubtful transfer stays closer to the observed TN.
The role classifier, projections, and weights remain fixed as $f_\theta$ learns.
Appendix~\ref{app:complete-trust} specifies their construction.

\paragraph{Protecting uncertain alarms.} Changing the TN comparison score alone does not prevent a plausible FP from being repeatedly pushed below it.
We assign each alarm a \emph{soft protection} weight $q_i^{\mathrm{prot}}\in[0,1]$, computed separately in each pool $\mathcal V\in\{\mathcal D_U,\mathcal C\}$.
Training-alarm evidence combines running auditor scores and proximity to the role boundary.
Target-alarm evidence also uses frozen precursor predictions, with transfer trust and allocation overlap moderating their influence.
Precursors initialize protection; the independent reference auditors below provide the distillation targets.

Let $m_{\mathcal V}$ be the protection mass assigned to pool $\mathcal V$.
An order-preserving allocator $\mathcal A_{m_{\mathcal V}}$ maps alarm evidence $e_i$ to $q_i^{\mathrm{prot}}=\sigma(e_i-\xi)$, where the shared offset $\xi$ is chosen so that $\sum_{i\in\mathcal V}q_i^{\mathrm{prot}}=m_{\mathcal V}$.
Here $\sigma(z)=(1+\exp(-z))^{-1}$ is the logistic function.
The mass is $m_{\mathcal V}=\lceil \widehat\rho_{\mathrm{tr}}|\mathcal V|\rceil$.
The training-only quantity $\widehat\rho_{\mathrm{tr}}$ measures TN-supported safe mass among alarms; it is not claimed to identify their global FP prevalence (Appendix~\ref{app:class-prior}).
The mass $m_{\mathcal V}$ is distinct from the human review budget $B$.

Let $\ell_{i,\mathcal V}$ be the comparison loss for alarm $i$, averaged over the transferred TN scores $s_p^{\mathrm{ref}}$.
With $\eta_{\mathcal C}\geq0$ denoting the relative use of the unlabeled target cohort, we train against
\begin{equation}
\mathcal L_{\mathrm{cmp},\mathcal V}
=\frac{\sum_{i\in\mathcal V}
 (1-q_i^{\mathrm{prot}})\ell_{i,\mathcal V}}
 {\sum_{i\in\mathcal V}(1-q_i^{\mathrm{prot}})},
\qquad
\mathcal J_{\mathrm{PU}}
=\frac{\mathcal L_{\mathrm{cmp},\mathcal D_U}
 +\eta_{\mathcal C}\mathcal L_{\mathrm{cmp},\mathcal C}}
 {1+\eta_{\mathcal C}}.
\label{eq:method:protected-comparison}
\end{equation}
Protection reduces comparison pressure; it does not assign an alarm a safe label.
The two pools retain separate updating memories.
A reference auditor minimizes $\mathcal L_{\mathrm{base}}+\mathcal J_{\mathrm{PU}}$.

\subsubsection{Reliability-Gated Rank Distillation}
\label{sec:method:rank-distillation}
\paragraph{Consensus positions.} Even with adapted supervision, independently fitted PU auditors can place uncertain alarms in different positions.
We use a frozen bank $\{f_{\phi_k}\}_{k=1}^{K}$ and transfer its ordering, rather than its raw score scale, into one student.
Let $\operatorname{Rank}(\mathbf a)$ denote ascending midpoint ranks normalized to $(0,1)$; equal values receive equal average ranks.
For reference $k$, define $r_i^{(k)} =\operatorname{Rank}((f_{\phi_k}(\mathbf h_j))_{j=1}^{N})_i$.
Their median and a fixed soft allocation are defined using $\operatorname{logit}(u)=\log(u/(1-u))$:
\begin{equation}
\bar r_i=\operatorname{median}_{1\leq k\leq K}r_i^{(k)},
\qquad
\mathbf q^{\mathrm{ref}}
=\mathcal A_{m_{\mathcal C}}
\bigl(\operatorname{logit}(\bar{\mathbf r})\bigr).
\label{eq:method:rank-consensus}
\end{equation}
The median limits the effect of an extreme reference position. $\mathbf q^{\mathrm{ref}}$ is fixed during student fitting and encodes review-order preference, not a safety label.

\paragraph{Pairwise transfer.} For target alarms $i\ne j$, let $w_{ij}^{\mathrm{dist}} =q_i^{\mathrm{ref}}(1-q_j^{\mathrm{ref}}) |\bar r_i-\bar r_j|$.
Let $s_i=f_\theta(\mathbf h_i)$ be the student score and $\mu\geq0$ the ranking margin.
The distillation loss is
\begin{equation}
\mathcal L_{\mathrm{dist}}
=\frac{\sum_{i\ne j}w_{ij}^{\mathrm{dist}}
 \log(1+\exp(\mu-s_i+s_j))}
 {\sum_{i\ne j}w_{ij}^{\mathrm{dist}}}.
\label{eq:method:rank-distillation}
\end{equation}
Both pair directions remain in the loss.
Their unequal weights favor the reference-preferred direction; the rank gap downweights nearly tied positions.
The gap is a measure of \emph{separation}, not pair-specific agreement among references.

\paragraph{One reliability gate.} Agreement alone cannot reveal a bias shared by all references.
We therefore use a cohort-level gate $\lambda_{\mathrm{dist}} =o_{\mathrm{ref}}(\bar\lambda^{\mathrm{tr}})^2$, where $o_{\mathrm{ref}}\in[0,1]$ is the soft-allocation overlap of a fixed reference pair and $\bar\lambda^{\mathrm{tr}}$ is mean TN transfer weight.
A lower overlap or weaker transfer reduces the influence of distillation.
The student minimizes
\begin{equation}
\mathcal L_{\mathrm{train}}
=\mathcal L_{\mathrm{base}}
 +\frac{\mathcal J_{\mathrm{PU}}
 +\lambda_{\mathrm{dist}}\mathcal L_{\mathrm{dist}}}
 {1+\lambda_{\mathrm{dist}}}.
\label{eq:method:student-objective}
\end{equation}
The base terms retain their weights; only PU comparison and distillation are blended.
Reference ranks and the gate stay fixed while the student's protection memory updates.
Stage~I returns one student, and keeps reference-rank dispersion for Stage~II (Appendix~\ref{app:rank-distillation}).

\subsection{Stage II: Consensus-Guided Structural Refinement}
\label{sec:method:stage-two}
The student captures a common ordering, but that ordering does not explicitly express how TN references support alarms at different granularities or how alarms relate within the target cohort.
We freeze the student and use those two kinds of structure to refine its review queue.
No student parameter is updated in this stage.

\paragraph{Hierarchical TN support.} Training role--group centers form a root--parent--leaf hierarchy.
When local TN support is sparse, the hierarchy allows a score to borrow support from a broader ancestor.
Holding the student score fixed, we form three \emph{score views} on the cohort: $\mathbf s^{\mathrm{local}}$ uses local support, $\mathbf s^{\mathrm{flat}}$ shrinks directly from leaf to root, and $\mathbf s^{\mathrm{hier}}$ shares support through the parent.
Their matched contrast defines
\begin{equation}
\mathbf s^{(0)}
=\mathbf s^{\mathrm{local}}
 +(\mathbf s^{\mathrm{hier}}-\mathbf s^{\mathrm{flat}}).
\label{eq:method:hierarchical-increment}
\end{equation}
The difference adds parent-level information without counting the shared student score twice.
Appendix~\ref{app:complete-geometry} defines the support and the three views.

\paragraph{Relations among target alarms.} The three views also suggest preferences between alarms.
We retain a cross-group directed pair $(i,j)$ only when the root view and at least one other support view agree that $i$ should precede $j$.
Let $\mathcal E$ be this edge set, $w_{ij}^{\mathrm{str}}\geq0$ its support weight, and $\delta_{ij}\geq0$ its desired separation.
A sigmoid transform of the ranks of $\mathbf s^{(0)}$ yields $\mathbf p\in(0,1)^N$.
We solve the following constrained problem, where $\operatorname{Ber}(u)$ denotes a Bernoulli distribution with mean $u$, $\operatorname{KL}$ is its Kullback--Leibler divergence, and $[z]_+=\max\{z,0\}$:
\begin{equation}
\begin{aligned}
\min_{\mathbf v\in\Omega(\mathbf p)}\quad&
\frac1N\sum_i
\operatorname{KL}\!\left(
\operatorname{Ber}(v_i)\,\|\,\operatorname{Ber}(p_i)\right)
 +\frac{\sum_{(i,j)\in\mathcal E}
w_{ij}^{\mathrm{str}}[\delta_{ij}-v_i+v_j]_+^2}
{\sum_{(i,j)\in\mathcal E}w_{ij}^{\mathrm{str}}},\\
\Omega(\mathbf p)
&=\{\mathbf v\in[0,1]^N:
\mathbf 1^\top\mathbf v=\mathbf 1^\top\mathbf p\}.
\end{aligned}
\label{eq:method:structural-objective}
\end{equation}
The first term keeps the solution near the hierarchy-supported initialization; the second penalizes violated relations.
The constraints retain its total soft mass.
With no supported relations, the solution is $\mathbf p$.
Let $\widehat{\mathbf v}$ be the finite solver output.
The structural ordering is formed with the endpoint-normalized ascending rank $\operatorname{Rank}_{01}$, which maps the smallest and largest positions to 0 and 1 (Appendix~\ref{app:rank-operators}):
\begin{equation}
\mathbf s^{\mathrm{str}}
=(1-\alpha)\operatorname{Rank}_{01}(\mathbf s^{(0)})
 +\alpha\operatorname{Rank}_{01}(\widehat{\mathbf v}),
\label{eq:method:structural-readout}
\end{equation}
where $\alpha\in[0,1]$ controls the contribution of cohort relations.
Appendices~\ref{app:rank-operators} and \ref{app:complete-graph} give the operators and solver details.

\paragraph{Agreement-guided readout.} We use the reference bank to estimate how stable the ranking of each alarm is.
Its rank spread and relative agreement target are
\begin{equation}
d_i=\max_{1\leq k\leq K}r_i^{(k)}
 -\min_{1\leq k\leq K}r_i^{(k)},
\qquad
c_i^\star=1-\operatorname{Rank}(\mathbf d)_i.
\label{eq:method:agreement-target}
\end{equation}
A ridge head predicts $c_i\in[0,1]$ from standardized frozen monitor features.
It is fitted once per cohort to $c_i^\star$, without alarm safety labels, and shared across student-seed repetitions.
Unlike the cohort-level distillation gate, $c_i$ controls an \emph{individual} alarm's final adjustment.

Let $\mathbf r^{\mathrm{stu}} =\operatorname{Rank}((f_\theta(\mathbf h_i))_{i=1}^{N})$ and $\mathbf r^{\mathrm{str}} =\operatorname{Rank}(\mathbf s^{\mathrm{str}})$.
For a fixed adjustment scale $\beta\geq0$, final review priority is
\begin{equation}
s_i^{\mathrm{out}}
=r_i^{\mathrm{str}}
 +\beta c_i(r_i^{\mathrm{str}}-r_i^{\mathrm{stu}}).
\label{eq:method:final-score}
\end{equation}
Predicted agreement determines how far to move along the structural correction; it is not itself evidence of safety.
The final scores need not lie in $[0,1]$ and are sorted in descending order.
Appendix~\ref{app:complete-agreement} specifies feature standardization, ridge fitting, and score handling.

The fitted cohort is scored with one student, the agreement head, and stored structural statistics.
A new cohort requires its own reference ranks and pool-dependent statistics; it does not require changing the frozen monitor's decisions.

\section{Experiments}
\label{sec:experiments}
\begingroup
\raggedbottom
\setlength{\textfloatsep}{6pt plus 1pt minus 1pt}
\setlength{\floatsep}{9pt plus 1pt minus 1pt}
\setlength{\intextsep}{4pt plus 1pt minus 1pt}
\setlength{\columnsep}{13pt}
\subsection{Experimental Setup}
\label{sec:exp-setup}

\paragraph{Protocol.}
We evaluate four dataset--monitor settings on ATBench~\citep{li2026atbench} and TraceSafe~\citep{chen2026tracesafe}, covering AgentDoG~\citep{liu2026agentdog}, Granite Guardian~\citep{padhi2025graniteguardian}, and StepGuard~\citep{zheng2026stepguard}.
Each setting provides 40 verified-safe non-alarms and unlabeled training alarms, with 510 target alarms in total.
Target labels remain sealed until all rankings are fixed, and all safety monitors remain frozen.

\paragraph{Baselines and metrics.}
We compare against eight PU methods: nnPU~\citep{kiryo2017nnpu}, DC-PU~\citep{li2025dcpu}, Dist-PU~\citep{zhao2022distpu}, Robust-PU~\citep{zhu2023robustpu}, PULDA~\citep{jiang2023pulda}, PUL-CPBF~\citep{li2024pulcpbf}, HolisticPU~\citep{wang2023holisticpu}, and feature-adapted NcPU~\citep{zhao2026ncpu}.
All methods use the same frozen features and data splits.
We report macro AUPRC/AUROC ($\uparrow$), AUGRC ($\downarrow$), and false alarms recovered under $5\%$, $10\%$, and $20\%$ review budgets.
Full implementation and evaluation details are in Appendix~\ref{app:training}.

\subsection{Ranking Performance and Review Yield}
\label{sec:main-results}

\textbf{\emph{Our method achieves macro AUPRC/AUROC/AUGRC of
$64.44/61.70/22.85$, improving over the strongest PU baseline by
$5.27/2.32/0.56$ favorable points and obtaining the strongest result across all
$96$ evaluated setting--seed--baseline comparisons.}}
\textbf{\emph{The advantage is particularly pronounced under limited review budgets:
at $5\%$, we recover $22.67$ false alarms versus PULDA's $17.00$
($+33.3\%$), while the gain remains substantial at $20\%$
($65.00$ versus $54.33$, $+19.6\%$).}}

\begin{wrapfigure}[15]{h}{0.45\linewidth}
\vspace{-17pt}
\centering
\includegraphics[width=\linewidth]{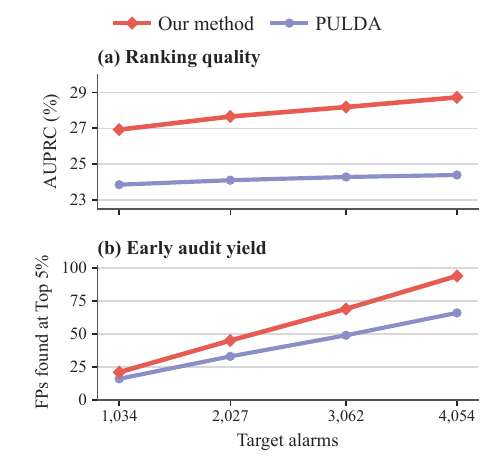}
\vspace{-25pt}
\caption{\textbf{External cohort scaling.}
Macro AUPRC and Top-$5\%$ false-alarm recovery.}
\label{fig:external-scaling}
\vspace{-19pt}
\end{wrapfigure}
\begin{table}[!ht]
\centering
\caption{\textbf{Ranking performance and review yield.} Upper: two ATBench settings.
Lower: four-setting macro mean${\pm\mathrm{std}}$ over seeds $\{13,37,73\}$ ($100\times$) and summed FP yield.
\textbf{Bold} and \underline{underlining} mark the best and second-best values.}
\label{tab:main-results}
\begingroup
\small
\setlength{\tabcolsep}{3pt}
\renewcommand{\arraystretch}{1.00}
\begin{tabularx}{\linewidth}{@{}l>{\centering\arraybackslash}X>{\centering\arraybackslash}X>{\centering\arraybackslash}X>{\centering\arraybackslash}X>{\centering\arraybackslash}X>{\centering\arraybackslash}X@{}}
\toprule
& \multicolumn{3}{c}{\shortstack{ATBench\\AgentDoG1.5-Qwen3.5-2B}}
& \multicolumn{3}{c}{\shortstack{ATBench\\Granite-Guardian-3.1-2B}} \\
\cmidrule(lr){2-4}\cmidrule(l){5-7}
Method (year) & AUPRC $\uparrow$ & AUROC $\uparrow$ & AUGRC $\downarrow$
& AUPRC $\uparrow$ & AUROC $\uparrow$ & AUGRC $\downarrow$ \\
\midrule

nnPU {\scriptsize(2017)}
& 61.75${\pm 0.09}$ & 65.94${\pm 0.15}$ & 19.66${\pm 0.04}$
& 60.73${\pm 0.24}$ & 50.97${\pm 0.35}$ & 21.09${\pm 0.09}$ \\

Dist-PU {\scriptsize(2022)}
& 70.04${\pm 2.76}$ & 70.00${\pm 3.57}$ & 18.65${\pm 0.89}$
& 68.16${\pm 1.12}$ & 57.81${\pm 1.57}$ & 19.43${\pm 0.38}$ \\

Robust-PU {\scriptsize(2023)}
& 62.02${\pm 0.55}$ & 66.75${\pm 0.71}$ & 19.46${\pm 0.18}$
& 60.24${\pm 0.19}$ & 50.78${\pm 0.14}$ & 21.14${\pm 0.03}$ \\

PULDA {\scriptsize(2023)}
& \underline{72.89}${\pm 0.92}$ & \underline{73.59}${\pm 0.51}$ & \underline{17.76}${\pm 0.13}$
& \underline{69.82}${\pm 0.65}$ & \underline{58.43}${\pm 1.17}$ & \underline{19.28}${\pm 0.29}$ \\

HolisticPU {\scriptsize(2023)}
& 61.85${\pm 1.18}$ & 65.77${\pm 1.14}$ & 19.71${\pm 0.28}$
& 57.97${\pm 0.54}$ & 48.06${\pm 0.53}$ & 21.80${\pm 0.13}$ \\

PUL-CPBF {\scriptsize(2024)}
& 61.95${\pm 2.06}$ & 64.70${\pm 2.09}$ & 19.97${\pm 0.52}$
& 61.56${\pm 2.24}$ & 51.28${\pm 4.26}$ & 21.02${\pm 1.04}$ \\

DC-PU {\scriptsize(2025)}
& 51.23${\pm 8.92}$ & 47.35${\pm 11.13}$ & 24.30${\pm 2.75}$
& 51.72${\pm 2.30}$ & 42.99${\pm 4.03}$ & 23.04${\pm 0.98}$ \\

NcPU {\scriptsize(2026)}
& 60.46${\pm 1.50}$ & 64.57${\pm 1.69}$ & 20.00${\pm 0.42}$
& 61.19${\pm 0.39}$ & 51.52${\pm 0.69}$ & 20.96${\pm 0.17}$ \\

\midrule
\rowcolor{black!7}
\textbf{Ours}
& \textbf{79.48}${\pm 0.65}$ & \textbf{76.67}${\pm 0.59}$ & \textbf{17.00}${\pm 0.15}$
& \textbf{73.91}${\pm 0.81}$ & \textbf{62.55}${\pm 0.86}$ & \textbf{18.27}${\pm 0.21}$ \\

\midrule
\addlinespace[4pt]
& \multicolumn{3}{c}{Average over all four combinations}
& \multicolumn{3}{c}{FPs found over all four combinations} \\
\cmidrule(lr){2-4}\cmidrule(l){5-7}
Method (year) & AUPRC $\uparrow$ & AUROC $\uparrow$ & AUGRC $\downarrow$
& Top 5\% & Top 10\% & Top 20\% \\
\midrule

nnPU {\scriptsize(2017)}
& 53.41${\pm 0.17}$ & 55.15${\pm 0.14}$ & 24.45${\pm 0.03}$
& 12.67 & 24.67 & 52.33 \\

Dist-PU {\scriptsize(2022)}
& 57.78${\pm 0.45}$ & 58.08${\pm 0.59}$ & 23.73${\pm 0.15}$
& 16.00 & 29.33 & \underline{54.33} \\

Robust-PU {\scriptsize(2023)}
& 53.14${\pm 0.17}$ & 54.99${\pm 0.16}$ & 24.49${\pm 0.04}$
& 12.00 & 24.00 & 50.00 \\

PULDA {\scriptsize(2023)}
& \underline{59.17}${\pm 0.25}$ & \underline{59.38}${\pm 0.40}$ & \underline{23.41}${\pm 0.10}$
& \underline{17.00} & \underline{31.00} & 53.33 \\

HolisticPU {\scriptsize(2023)}
& 53.55${\pm 0.43}$ & 55.28${\pm 0.38}$ & 24.42${\pm 0.09}$
& 13.33 & 25.00 & 52.33 \\

PUL-CPBF {\scriptsize(2024)}
& 52.96${\pm 1.19}$ & 54.03${\pm 1.98}$ & 24.74${\pm 0.49}$
& 11.00 & 25.00 & 48.33 \\

DC-PU {\scriptsize(2025)}
& 47.46${\pm 2.78}$ & 47.50${\pm 4.21}$ & 26.26${\pm 1.08}$
& 11.33 & 19.33 & 43.67 \\

NcPU {\scriptsize(2026)}
& 53.71${\pm 0.28}$ & 55.59${\pm 0.50}$ & 24.35${\pm 0.12}$
& 14.67 & 24.67 & 51.67 \\

\midrule
\rowcolor{black!7}
\textbf{Ours}
& \textbf{64.44}${\pm 0.21}$ & \textbf{61.70}${\pm 0.20}$ & \textbf{22.85}${\pm 0.05}$
& \textbf{22.67} & \textbf{37.00} & \textbf{65.00} \\

\bottomrule
\end{tabularx}
\endgroup

\par\smallskip
{\footnotesize\raggedright
The 5\%, 10\%, and 20\% budgets review 27, 52, and 104 alarms, respectively,
across 510 target alarms containing 239 hidden FPs.\par}
\end{table}

\subsection{Large-scale External Validation}
\label{sec:external-validation}

We further evaluate on Agent-SafetyBench~\citep{zhang2024agentsafetybench},
yielding 4,054 held-out alarms from 1,400 evaluation cases and two frozen monitors.
\textbf{\emph{Our method improves PULDA from $24.39$ to $28.72$ AUPRC and increases
Top-$5\%$ recovery from $66$ to $94$ false alarms ($+42.4\%$).}}
The AUPRC improvement is supported by a case-level $95\%$ bootstrap interval
of $[3.12,5.58]$.
Even the $K=1$ variant outperforms a three-model PULDA ensemble
($26.15$ versus $25.42$), while the full method maintains its advantage as
the target cohort grows from 1,034 to 4,054 alarms.
This consistent scaling indicates that the improvement transfers beyond the
smaller primary evaluation to substantially larger audit pools.

\begin{table}[h]
\centering
\caption{\textbf{Large-scale external validation on Agent-SafetyBench.}
Ranking metrics are two-monitor macro scores ($100\times$); FP columns report
recovered false alarms at each review budget.}
\label{tab:external-main}
\small
\setlength{\tabcolsep}{4pt}
\begin{tabular}{lcccccc}
\toprule
Method & AUPRC $\uparrow$ & AUROC $\uparrow$ & AUGRC $\downarrow$
& FP@5\% $\uparrow$ & FP@10\% $\uparrow$ & FP@20\% $\uparrow$ \\
\midrule
Dist-PU & 24.16 & 57.20 & 40.97 & 66 & 110 & 165 \\
PULDA & 24.39 & 58.57 & 40.74 & 66 & 106 & 186 \\
PULDA ensemble & 25.42 & 59.45 & 40.21 & 71 & 114 & 195 \\
Our method ($K=1$) & 26.15 & 60.32 & 39.68 & 76 & 125 & 208 \\
\textbf{Our method ($K=3$)}
& \textbf{28.72} & \textbf{63.45} & \textbf{37.82}
& \textbf{94} & \textbf{152} & \textbf{241} \\
\bottomrule
\end{tabular}
\end{table}

\subsection{Direct LLM Re-judgment}
\label{sec:llm-judge}

We compare against a zero-shot Qwen2.5-7B-Instruct judge that independently
re-evaluates each policy, task, and trajectory without access to labels,
safe references, or monitor scores.
Although direct semantic re-judgment improves over the native monitor order,
our method achieves substantially higher AUPRC ($64.44$ versus $57.78$)
and recovers $4.67$, $9.00$, and $15.00$ more false alarms at the
$5\%$, $10\%$, and $20\%$ budgets, respectively
(\cref{fig:main-controls}(a,d)).
\textbf{\emph{This gap shows that cohort-level PU supervision provides information beyond
independent semantic re-evaluation of individual alarms.}}

\subsection{Ablation and Robustness}
\label{sec:components}

\textbf{\emph{Each component contributes consistently:
trust-aware supervision, rank distillation, and structural refinement increase
AUPRC from $58.42$ to $60.35$, $61.92$, and $64.44$, respectively,
while improving all three ranking metrics at every stage.}}

% Results supplied by the author on 2026-09-19; no rerun or independent validation.
\begingroup
\centering
\captionof{table}{\textbf{Cumulative component ablation.}
Rows add one module; metrics use a $100\times$ scale.
\textbf{Bold}/\underline{underlining}: best/second-best.}
\label{tab:components}
\vspace{-10pt}
\begingroup
\small
\setlength{\tabcolsep}{5pt}
\renewcommand{\arraystretch}{1.08}
\begin{tabularx}{\linewidth}{@{}cX!{\vrule width 0.35pt}ccc@{}}
\toprule
ID & Configuration & AUPRC $\uparrow$ & AUROC $\uparrow$ & AUGRC $\downarrow$ \\
\midrule
(1) & Base PU student & 58.42 & 57.80 & 23.650 \\
(2) & + Trust-aware PU supervision & 60.35 & 59.92 & 23.210 \\
(3) & + Reliability-gated rank distillation & \underline{61.92} & \underline{61.34} & \underline{22.955} \\
\midrule
\rowcolor{black!8}(4) & \textbf{+ Structural queue refinement (Ours)} & \textbf{64.44} & \textbf{61.70} & \textbf{22.846} \\
\bottomrule
\end{tabularx}
\endgroup
\par\smallskip
\begin{minipage}{\linewidth}
\noindent Each component improves all three metrics; together they add $6.02$ AUPRC and $3.90$ AUROC points and reduce AUGRC by $0.804$.
\end{minipage}
\par\endgroup

\begin{table}[h]
\centering
\caption{\textbf{Local LLM re-judgment control.} Four-setting macro ranking and summed FP yield; the Judge uses its frozen bidirectional score.
\textbf{Bold}/\underline{underlining}: best/second-best.}
\label{tab:llm-judge-main}
\begingroup
\footnotesize
\setlength{\tabcolsep}{4.2pt}
\renewcommand{\arraystretch}{1.08}
\begin{tabular*}{\linewidth}{@{\extracolsep{\fill}}lcccccc@{}}
\toprule
& \multicolumn{3}{c}{Macro ranking ($100\times$)}
& \multicolumn{3}{c}{FPs recovered} \\
\cmidrule(lr){2-4}\cmidrule(l){5-7}
Method & AUPRC $\uparrow$ & AUROC $\uparrow$ & AUGRC $\downarrow$
& Top 5\% & Top 10\% & Top 20\% \\
\midrule
Native monitor score & 53.39 & 55.21 & 24.71 & 12.00 & 24.00 & 50.00 \\
PULDA & \underline{59.17} & \underline{59.38} & \underline{23.41}
& 17.00 & \underline{31.00} & \underline{53.33} \\
Qwen2.5-7B judge & 57.78 & 57.17 & 23.97
& \underline{18.00} & 28.00 & 50.00 \\
\midrule
\rowcolor{black!7}\textbf{Our method} & \textbf{64.44} & \textbf{61.70}
& \textbf{22.85} & \textbf{22.67} & \textbf{37.00} & \textbf{65.00} \\
\bottomrule
\end{tabular*}
\endgroup
\end{table}

\begin{figure}[h]
\centering
\includegraphics[width=\linewidth]{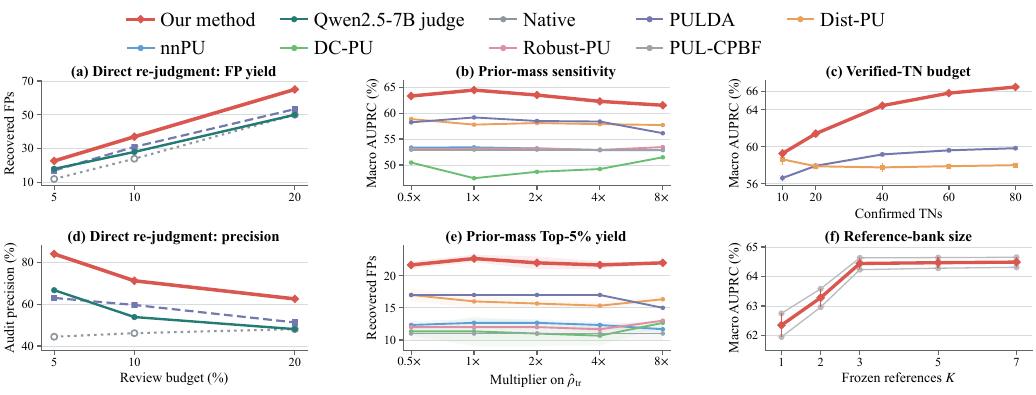}
\caption{\textbf{Controls and sensitivity analyses.}
(a,d) Direct LLM re-judgment;
(b,e) prior-mass sensitivity;
(c) verified-safe supervision;
(f) reference-bank size.}
\label{fig:main-controls}
\end{figure}

The improvement is not explained by simple ensembling:
our single-reference variant reaches $62.35$ AUPRC, exceeding the
three-model PULDA ensemble at $60.24$.
The method is also robust to its training-only prior-mass estimate:
across a $0.5$--$8\times$ range, AUPRC remains $61.52$--$64.44$
and Top-$5\%$ recovery remains $21.67$--$22.67$.
Performance further improves with additional verified-safe supervision,
rising from $59.28$ AUPRC with 10 references to $66.45$ with 80.
At the default 40-reference setting, $K=3$ captures nearly all of the observed
benefit from increasing the reference bank
(\cref{fig:main-controls}(b--f)).
Together, these results show that the gains persist across both supervision
levels and reference-bank configurations.

\subsection{Deployment Cost}
\label{sec:deployment-cost}

With cached features, the default $K=3$ configuration requires
$162.86$ seconds of offline fitting on one NVIDIA H200.
Deployment uses a single student scorer without reference-model ensembling,
keeping the final auditing pipeline compact despite the richer training procedure.
Detailed cost analysis is provided in Appendix~\ref{app:cost-and-deployment}.

\section{Conclusion}

In this work, we study false-alarm auditing for agent safety monitors under monitor-induced selection, where verified-safe non-alarms may poorly represent the hidden safe alarms of interest.
We develop a two-stage PU ranking framework that first adapts verified-safe supervision toward the alarm domain and reduces excessive negative pressure on plausible false alarms, then stabilizes ranking preferences across independently trained PU models and further exploits hierarchical safe-reference support and relations among alarms.
Across mainstream safety monitors, our method improves macro AUPRC over eight PU baselines by $5.27$--$16.98$ absolute percentage points and recovers $33.3\%$ more false alarms than PULDA at a $5\%$ review budget, with consistent gains on a larger external benchmark of $4{,}054$ held-out alarms.
The framework requires no alarm safety labels for fitting and leaves the monitor decisions unchanged.
\FloatBarrier

\clearpage
\endgroup
\section*{AI Use Statement}
Generative AI tools were used to assist with language editing, manuscript organization, and feedback on presentation. All research decisions, experimental designs, implementations, analyses, and reported results were reviewed and verified by the authors, who take full responsibility for the final content.
\section*{Ethics Statement}
This work prioritizes alarms for human review rather than automatically suppressing them. A high-ranked alarm is not certified safe, and the original monitor decisions remain in force. Deployment should keep reviewers in the loop and respect the privacy and access conditions of the underlying agent traces.

\section*{Reproducibility Statement}
The main text and appendix specify the method, evaluation protocol, and fixed
training configuration.
The code repository contains the core implementation, fixed configurations,
sample manifests, frozen predictions, implementation tests, and scripts for
regenerating the reported results.
Benchmark data and checkpoints follow their original distribution licenses.

\bibliography{references}
\bibliographystyle{iclr2027_conference}
\appendix
% ======================================================================
% Appendix: implementation, controls, and extended evaluation
% ======================================================================

\section{Training Definitions and Implementation}
\label{app:training}

We provide the implementation details required to reproduce the reported system.
All training diagnostics use only verified-safe TNs and unlabeled training alarms,
while target safety labels remain sealed until all ranking scores are fixed.
The same configuration is used across all reported dataset--monitor settings.

% ----------------------------------------------------------------------
\subsection{Rank Operators}
\label{app:rank-operators}

For $\mathbf a\in\mathbb R^n$, let
$\operatorname{arank}(\mathbf a)$ denote ascending ranks with midpoint tie
handling. Reference models use
\begin{equation}
\operatorname{Rank}(\mathbf a)_i
=
\frac{\operatorname{arank}(\mathbf a)_i-\tfrac12}{n}.
\label{eq:app:midpoint-ranks}
\end{equation}
Structural refinement uses
$\operatorname{Rank}_{01}(\mathbf a)_i
=(\operatorname{arank}(\mathbf a)_i-1)/(n-1)$ for $n>1$.
Sample ID resolves remaining ties.
We use
$\sigma(z)=(1+\exp(-z))^{-1}$ and
$\operatorname{logit}(u)=\log(u/(1-u))$.

% ----------------------------------------------------------------------
\subsection{TN Transfer and Soft Protection}
\label{app:complete-trust}
\label{app:complete-memory}

\paragraph{Transfer diagnostics.}
A five-fold, isolation-group-disjoint logistic role classifier is fitted on
verified TNs and training alarms using normalized monitor and task
representations.
For TN $\tau_p$, let $\ell_p$ be its held-out role logit,
$A_{\mathrm{role}}$ the task-block AUC, and
$\mathbf t_p$ its normalized task representation.
With role-classifier parameters
$\varphi=(\mathbf w_h,\mathbf w_t,b)$, define
\[
\mathbf a=\frac{\mathbf w_h}{\|\mathbf w_h\|_2},
\qquad
c_p=-\frac{\mathbf w_t^\top\mathbf t_p+b}{\|\mathbf w_h\|_2},
\qquad
\mathbf x_p=\frac{\mathbf h_p}{\|\mathbf h_p\|_2}.
\]

The nearest role-boundary representation preserving task features and hidden
norm is
\begin{equation}
\widetilde{\mathbf h}_p
=
\|\mathbf h_p\|_2
\left[
c_p\mathbf a+
\sqrt{1-c_p^2}
\frac{\mathbf x_p-(\mathbf a^\top\mathbf x_p)\mathbf a}
{\|\mathbf x_p-(\mathbf a^\top\mathbf x_p)\mathbf a\|_2}
\right].
\label{eq:app:boundary-projection}
\end{equation}

Let
$d_p=\|\widetilde{\mathbf h}_p/\|\mathbf h_p\|_2-\mathbf x_p\|_2$.
Transfer support is
\begin{equation}
u_p^{\mathrm{role}}
=
\exp\!\left[
-\left(\frac{1-A_{\mathrm{role}}}{0.02}\right)^8
-|\ell_p|^8
\right],
\qquad
u_p^{\mathrm{geom}}
=
\exp\!\left[
-\left(\frac{1-A_{\mathrm{role}}}{0.02}\right)^8
-\left(\frac{d_p}{0.12}\right)^8
\right].
\end{equation}

With
$c_{75}=\exp[-Q_{0.75}(|\ell|)^8]$ computed from training TNs,
\begin{equation}
\lambda_p^{\mathrm{tr}}
=
c_{75}u_p^{\mathrm{role}}
+
(1-c_{75})
\min(u_p^{\mathrm{role}},u_p^{\mathrm{geom}})
\in[0,1].
\label{eq:app:transfer-trust}
\end{equation}
All transfer quantities are fixed before PU fitting.

\paragraph{Soft protection.}
Each alarm receives evidence $e_i$ from running ranking information and
role-boundary proximity.
The order-preserving allocator produces
\begin{equation}
q_i^{\mathrm{prot}}=\sigma(e_i-\xi),
\qquad
\sum_{i\in\mathcal V}q_i^{\mathrm{prot}}=m_{\mathcal V},
\label{eq:complete-allocation}
\end{equation}
where $\xi$ is obtained by bisection.
Training and target pools maintain separate memories.
Protection adjusts comparison strength while preserving the original
unlabeled status of every alarm.

\paragraph{Task-conditioned comparisons.}
For normalized task representations $\mathbf t_p,\mathbf t_i$, define
\[
D_{pi}=\max\{0,2-2\mathbf t_p^\top\mathbf t_i\},
\qquad
\kappa_{pi}=\exp[-D_{pi}/(2h_{\mathrm{task}})],
\]
where $h_{\mathrm{task}}$ is the median positive training TN--alarm distance.
For TN minibatch $\mathcal B_P$, the comparison term is
\begin{equation}
\ell_{i,\mathcal V}
=
\sum_{p\in\mathcal B_P}
\nu_{pi,\mathcal V}
\log(1+\exp(\mu-s_p^{\mathrm{ref}}+s_i)).
\label{eq:app:alarm-comparison}
\end{equation}
These weights redistribute comparison pressure without introducing alarm labels.

% ----------------------------------------------------------------------
\subsection{PU Objective and Reference Fits}
\label{app:complete-base}

Every fitted scorer uses the common objective
\begin{equation}
\mathcal L_{\mathrm{base}}
=
\mathcal L_{\mathrm{align}}
+
\lambda_{\mathrm{mix}}(t)\mathcal L_{\mathrm{mix}}.
\label{eq:app:base-objective}
\end{equation}

For TN and alarm minibatches, let $\bar p_P$ and $\bar p_U$ be their mean
predicted safety probabilities.
With
$\zeta_T(z)=T^{-1}\log(1+\exp(Tz))$ and
$D_T(p,\rho)=\zeta_T(p-\rho)+\zeta_T(\rho-p)$,
the alignment term is
\begin{equation}
\mathcal L_{\mathrm{align}}
=
2\rho(1-\bar p_P)
+
\begin{cases}
D_T(\bar p_U,\rho),
& \text{first alarm minibatch},\\
D_T(\widetilde p_U,\rho)/(1-a_U),
& \text{otherwise},
\end{cases}
\label{eq:app:alignment-loss}
\end{equation}
where
$\widetilde p_U
=a_U\widetilde p_U^{\mathrm{prev}}+(1-a_U)\bar p_U$
and $\rho=\widehat\rho_{\mathrm{tr}}$.

For group-structured mixing, let
$\mathbf c_i$ be the role--isolation-group centroid,
$\boldsymbol\mu_{\chi_i}$ the corresponding role mean,
$\mathbf b_i=\mathbf c_i-\boldsymbol\mu_{\chi_i}$, and
$\mathbf u_i=\mathbf h_i-\mathbf c_i$.
For a cross-role pair $(i,j)$,
\begin{equation}
\begin{aligned}
\mathbf h_i^{\mathrm{mix}}
={}&
\xi_i\boldsymbol\mu_{\chi_i}
+(1-\xi_i)\boldsymbol\mu_{\chi_j}
+\sqrt{\xi_i}\mathbf b_i
+\sqrt{1-\xi_i}\mathbf b_j \\
&+\xi_i\mathbf u_i+(1-\xi_i)\mathbf u_j .
\end{aligned}
\label{eq:app:mixed-representation}
\end{equation}
Same-role pairs use standard interpolation.

With detached soft targets $\widehat y_i$ and
$p_i^{\mathrm{mix}}
=\sigma(\operatorname{clip}(f_\theta(\mathbf h_i^{\mathrm{mix}}),-10,10))$,
\begin{equation}
\mathcal L_{\mathrm{mix}}
=
\frac1{|\mathcal B|}
\sum_{i\in\mathcal B}
\left[
\xi_i\operatorname{BCE}(p_i^{\mathrm{mix}},\widehat y_i)
+
(1-\xi_i)
\operatorname{BCE}(p_i^{\mathrm{mix}},\widehat y_{\varpi(i)})
\right].
\label{eq:app:mixture-loss}
\end{equation}

% ----------------------------------------------------------------------
\subsection{Reference Ranks and Student Distillation}
\label{app:rank-distillation}

Reference models use identical data, architecture, and PU objectives with
independent initialization.
Their frozen outputs are converted to normalized ranks, whose coordinatewise
median yields $\bar{\mathbf r}$.

The fixed reference allocation is
\[
\mathbf q^{\mathrm{ref}}
=
\mathcal A_{m_{\mathcal C}}
(\operatorname{logit}(\bar{\mathbf r})).
\]
For equal-mass allocations,
\[
o(\mathbf q,\mathbf q')
=
\frac{\sum_i\min(q_i,q_i')}{m_{\mathcal C}},
\]
and the cohort-level distillation weight is
\[
\lambda_{\mathrm{dist}}
=
o_{\mathrm{ref}}
(\bar\lambda^{\mathrm{tr}})^2.
\]
Reference ranks, pair weights, and the gate remain fixed while the student
is optimized, so the reference bank contributes relative ordering information
throughout distillation.

% ----------------------------------------------------------------------
\subsection{Optimization and Fixed Settings}
\label{app:rmd-relocations}

All scorers use a two-hidden-layer MLP with width $1000$.
Adam uses learning rate $10^{-3}$, weight decay $10^{-4}$, and gradient
clipping at $5$.
Each fit runs for $400$ epochs with TN/alarm minibatches of size $64$,
with the first $200$ epochs forming the common warm-up.
All reported dataset--monitor settings use the same optimization schedule
and hyperparameters; seeds $13$, $37$, and $73$ differ only in initialization
and minibatch order.

\begin{table}[!ht]
\centering
\caption{\textbf{Fixed implementation settings.} Values are shared across all four dataset--monitor cells and all seeds.}
\label{tab:hyperparameters}
\begingroup
\small
\setlength{\tabcolsep}{4pt}
\renewcommand{\arraystretch}{1.04}
\begin{tabularx}{\linewidth}{@{}lX@{}}
\toprule
Component & Setting \\
\midrule
PU scorer & two hidden layers; width $1000$ each; Adam, learning rate $10^{-3}$, weight decay $10^{-4}$; gradient clip $5$ \\
Training & $400$ epochs; TN/alarm batches $64/64$; $\max\{1,\lceil |\mathcal D_U|/64\rceil\}$ updates per epoch; $200$-epoch warm-up \\
Ranking and mixing & margin $0.6$; post-warm-up $\lambda_{\mathrm{mix}}=4.2$; $\operatorname{Beta}(11,11)$ mixing; protection-memory EMA $0.5$ \\
Distribution alignment & smoothing temperature $T=3.5$; alarm-mean EMA coefficient $a_U=0.85$ \\
Role and transfer & five group-disjoint role folds; logistic $C=1$; L-BFGS tolerance $10^{-10}$; role scale $1$; displacement scale $0.12$ \\
Reference bank & $K=3$; midpoint ranks; coordinatewise median; protection mass $\lceil\widehat\rho_{\mathrm{tr}}|\mathcal V|\rceil$ \\
Structural readout & $16$ PCA components; root/parent/leaf branching $1/2/2$; $50$ projected iterations; solver-rank blend $0.25$ \\
Agreement head & target standardization; constant-feature threshold $10^{-10}$; Ridge strength $100$; output clip $[0,1]$; final coefficient $0.20$ \\
Evaluation & seeds $13,37,73$; checkpoint epoch $400$; ascending sample ID resolves score ties \\
\bottomrule
\end{tabularx}
\endgroup
\end{table}

\FloatBarrier

% ----------------------------------------------------------------------
\subsection{TN-Supported Mass and Sensitivity}
\label{app:class-prior}

The training-only quantity $\widehat\rho_{\mathrm{tr}}$ controls the amount
of alarm mass eligible for soft protection rather than estimating global
false-alarm prevalence.

For a TN-like upper tail
$\mathcal H_t=\{\tau:g_\varphi(\tau)\ge t\}$,
\begin{equation}
R_t
=
\frac{\Pr_{\mathcal D_U}(\mathcal H_t)}
{\Pr_{\mathcal D_P}(\mathcal H_t)}
=
\pi_U\kappa_t+(1-\pi_U)\delta_t,
\label{eq:tail-ratio-semantics}
\end{equation}
which motivates the observed tail ratio as a training-only support statistic.

Isolation groups are split into three folds and the statistic is computed from
held-out role-classifier scores.
A single fixed construction is used across all reported settings before target
labels are accessed.

\begin{table}[!ht]
\centering
\caption{\textbf{Training-only TN-supported mass.}
Values are computed without alarm safety labels.}
\label{tab:prior-values}
\small
\setlength{\tabcolsep}{5pt}
\begin{tabular}{llc}
\toprule
Dataset & Monitor & $\widehat\rho_{\mathrm{tr}}$ \\
\midrule
ATBench & AgentDoG1.5-Qwen3.5-2B & 0.02381 \\
ATBench & Granite-Guardian-3.1-2B & 0.02368 \\
TraceSafe & AgentDoG-Qwen3-4B & 0.01000 \\
TraceSafe & StepGuard-4B & 0.01266 \\
\bottomrule
\end{tabular}
\end{table}

We evaluate
$\{0.5,1,2,4,8\}\widehat\rho_{\mathrm{tr}}$
for all applicable methods.
Our method remains strongest throughout the sweep, with AUPRC of
$61.52$--$64.44$ and Top-$5\%$ recovery of $21.67$--$22.67$,
compared with the strongest baseline operating points of $59.17$ and
$17.00$, respectively.

\begin{table}[!ht]
\centering
\caption{\textbf{Prior-mass sensitivity.}
Four-setting macro AUPRC under multipliers of the training-only
TN-supported mass.}
\label{tab:prior-sensitivity}
\small
\setlength{\tabcolsep}{5pt}
\begin{tabular}{lccccc}
\toprule
Method & $0.5\times$ & $1\times$ & $2\times$ & $4\times$ & $8\times$ \\
\midrule
nnPU       & 53.36 & 53.41 & 53.18 & 52.86 & 52.87 \\
DC-PU      & 50.48 & 47.46 & 48.69 & 49.22 & 51.49 \\
Dist-PU    & 58.86 & 57.78 & 58.09 & 57.88 & 57.71 \\
Robust-PU  & 53.04 & 53.14 & 53.21 & 52.91 & 53.45 \\
PUL-CPBF   & 52.96 & 52.96 & 52.96 & 52.96 & 52.96 \\
PULDA      & 58.23 & 59.17 & 58.48 & 58.38 & 56.13 \\
\textbf{Ours}
            & \textbf{63.30} & \textbf{64.44} & \textbf{63.49}
            & \textbf{62.27} & \textbf{61.52} \\
\bottomrule
\end{tabular}
\end{table}

\begin{center}
\captionof{table}{\textbf{Evaluation cells and fixed protocol.}
Training alarms are unlabeled during fitting; target FP counts are used only
for evaluation. Each cell contains 40 verified TNs.}
\label{tab:protocol-details}
\begingroup
\small
\setlength{\tabcolsep}{4pt}
\renewcommand{\arraystretch}{1.08}
\begin{tabularx}{\linewidth}{@{}XXccc@{}}
\toprule
Dataset & Monitor & Training alarms & Target alarms & Target FPs \\
\midrule
ATBench & AgentDoG1.5-Qwen3.5-2B & 249 & 56 & 30 \\
ATBench & Granite-Guardian-3.1-2B & 300 & 76 & 44 \\
TraceSafe & AgentDoG-Qwen3-4B & 949 & 269 & 122 \\
TraceSafe & StepGuard-4B & 504 & 109 & 43 \\
\midrule
\multicolumn{2}{@{}l}{Total target pool} & --- & 510 & 239 \\
\bottomrule
\end{tabularx}
\endgroup
\end{center}

Macro AUPRC, AUROC, and AUGRC give equal weight to the four settings.
At review budget $b$, we inspect $\lceil bN_c\rceil$ alarms per setting,
sum recovered FPs, and average over seeds.
Checkpoints are fixed at epoch $400$.

% ======================================================================
\section{Structural Queue Refinement}
\label{app:refinement}

Structural refinement is a frozen-student readout using training-derived TN
support, unlabeled target representations, and reference-rank consensus.

% ----------------------------------------------------------------------
\subsection{Hierarchical TN Support}
\label{app:complete-geometry}

Training role--group centers define a root--parent--leaf hierarchy.
When local TN support is weak, support shrinks toward the nearest reliable
ancestor:
\begin{equation}
u_g
=
\rho_g u_g^{\mathrm{local}}
+
(1-\rho_g)u_{\operatorname{par}(g)},
\qquad
\rho_g\in[0,1].
\label{eq:app:hierarchical-support}
\end{equation}
The reliability $\rho_g$ depends only on training support and geometry.
The local, flat, and hierarchical views share the same student contribution,
so their contrast isolates additional hierarchical support.

% ----------------------------------------------------------------------
\subsection{Consensus-Supported Relations}
\label{app:complete-graph}

Cross-group alarm relations are retained only when the root view and at least
one more specific support view agree on direction.
Starting from the hierarchy-supported initialization, we solve
\begin{equation}
\begin{aligned}
\min_{\mathbf v\in\Omega(\mathbf p)}
\quad&
\frac1N\sum_i
\operatorname{KL}\!\left(
\operatorname{Ber}(v_i)\|
\operatorname{Ber}(p_i)
\right)\\
&+
\frac{
\sum_{(i,j)\in\mathcal E}
w_{ij}^{\mathrm{str}}
[\delta_{ij}-v_i+v_j]_+^2
}{
\sum_{(i,j)\in\mathcal E}w_{ij}^{\mathrm{str}}
},
\\[-2pt]
\Omega(\mathbf p)
&=
\left\{
\mathbf v\in[0,1]^N:
\mathbf1^\top\mathbf v=\mathbf1^\top\mathbf p
\right\}.
\end{aligned}
\label{eq:app:structural-objective}
\end{equation}
We use $50$ projected iterations with monotone backtracking.
The student remains frozen throughout refinement.

% ----------------------------------------------------------------------
\subsection{Reference Consensus and Final Readout}
\label{app:complete-agreement}

Reference dispersion for alarm $i$ is
$d_i=\max_k r_i^{(k)}-\min_k r_i^{(k)}$, with relative consensus target
\begin{equation}
c_i^\star
=
1-\operatorname{Rank}(\mathbf d)_i.
\label{eq:app:agreement-target}
\end{equation}
A ridge head predicts this target from standardized frozen monitor
representations.
It is fitted once per target cohort using reference ranks only, without alarm
safety labels.

Predicted agreement scales the final structural displacement.
The same readout coefficient is used across all reported settings.
Deployment requires only the frozen student and stored cohort statistics,
without reference-model ensembling.

% ======================================================================
\section{Data Provenance and Experimental Controls}
\label{app:cost-and-deployment}
\begingroup
\small

% ----------------------------------------------------------------------
\subsection{Benchmarks and Monitor Identifiers}

ATBench follows \citet{li2026atbench};
AgentDoG follows \citet{liu2026agentdog};
TraceSafe refers to TraceSafe-Bench~\citep{chen2026tracesafe};
Granite Guardian follows \citet{padhi2025graniteguardian};
and StepGuard follows \citet{zheng2026stepguard}.
\Cref{tab:protocol-details} lists the exact benchmark--monitor pairings and
sample counts.

% ----------------------------------------------------------------------
\subsection{Computational Cost}

With cached monitor and task features, the default $K=3$ configuration
requires $162.86$ seconds of offline cohort-level fitting on one NVIDIA H200.
After fitting, deployment uses one student scorer together with lightweight
stored cohort statistics, without reference-model ensembling.

\begin{table}[!ht]
\centering
\caption{\textbf{Resource usage for the default $K=3$ configuration.}
Measurements start from cached monitor and task features on one NVIDIA H200.}
\label{tab:resource-cost}
\small
\setlength{\tabcolsep}{7pt}
\begin{tabular}{lc}
\toprule
Resource & $K=3$ \\
\midrule
Total elapsed time (s) & 162.86 \\
Peak GPU allocation (MiB) & 138.93 \\
Peak GPU reservation (MiB) & 164.00 \\
Peak process memory (MiB) & 2223.91 \\
\bottomrule
\end{tabular}
\end{table}

% ----------------------------------------------------------------------
\subsection{Local LLM-as-a-Judge Control}
\label{app:llm-judge}

We evaluate Qwen2.5-7B-Instruct as a zero-shot semantic re-judgment baseline.
For each target alarm, the model receives the safety policy, task instruction,
and serialized trajectory, but no target label, verified-safe reference,
target-set statistic, or monitor confidence.
The prompt and bidirectional scoring rule are fixed before labels are opened.

Direct re-judgment reaches $57.78$ macro AUPRC, while our method reaches
$64.44$ and recovers an additional $4.67$, $9.00$, and $15.00$ false alarms
at the $5\%$, $10\%$, and $20\%$ review budgets, respectively.

% ----------------------------------------------------------------------
\subsection{Deployment Protocol}

The method is designed for periodic batch auditing.
Deployment uses the frozen student and stored structural statistics, with
cohort-level statistics constructed using the same fixed procedure for each
audit batch.

\paragraph{Reproducibility.}
The supplement contains fixed configurations, sample manifests,
frozen predictions, implementation tests, and table-regeneration scripts.
Target labels are accessed only after prediction outputs are fixed.

\endgroup

% ======================================================================
\section{Additional Experimental Controls}
\label{app:additional-controls}

% ----------------------------------------------------------------------
\subsection{Selection and Representation Controls}

\begin{table}[!t]
\centering
\caption{\textbf{Selection-bias and representation controls.} Ranking metrics are macro-averaged over the same four target cells ($100\times$); yield is the summed number of recovered FPs.
SAR-EM uses the official selected-at-random implementation~\citep{bekker2018sar}.
Deterministic representation controls use only the 40 training TNs and frozen monitor representations.
PULDA and our method are the three-seed results from \cref{tab:main-results}.}
\label{tab:selection-controls}
\begingroup
\scriptsize
\setlength{\tabcolsep}{3.2pt}
\renewcommand{\arraystretch}{1.02}
\begin{tabularx}{\linewidth}{@{}l>{\centering\arraybackslash}X>{\centering\arraybackslash}X>{\centering\arraybackslash}Xccc@{}}
\toprule
Method & AUPRC $\uparrow$ & AUROC $\uparrow$ & AUGRC $\downarrow$ & Top 5\% & Top 10\% & Top 20\% \\
\midrule
Native monitor score & --- & --- & --- & --- & --- & 50.00 \\
TN centroid distance & 55.44 & 57.98 & 23.75 & 13.00 & 28.00 & 53.00 \\
$5$-NN to TNs & 54.69 & 57.74 & 23.82 & 14.00 & 27.00 & 50.00 \\
One-class SVM~\citep{scholkopf2001support} & 57.74 & 59.30 & 23.43 & 14.00 & 29.00 & 52.00 \\
TN PCA support & 54.97 & 57.55 & 23.86 & 14.00 & 29.00 & 51.00 \\
Role-propensity score & 53.89 & 55.93 & 24.26 & 13.00 & 26.00 & 53.00 \\
SAR-EM (role propensity) & 54.11 & 54.51 & 24.63 & 12.00 & 26.00 & 52.00 \\
PULDA & \underline{59.17} & \underline{59.38} & \underline{23.41} & \underline{17.00} & \underline{31.00} & \underline{53.33} \\
\rowcolor{black!7}\textbf{Ours} & \textbf{64.44} & \textbf{61.70} & \textbf{22.85} & \textbf{22.67} & \textbf{37.00} & \textbf{65.00} \\
\bottomrule
\end{tabularx}
\endgroup
\par\smallskip
{\footnotesize\raggedright Native outputs were archived consistently for all four cells at the $20\%$ cutoff; unavailable full-score entries are left blank rather than reconstructed. SAR-EM and the representation controls are deterministic fits, so no seed dispersion is attached.\par}
\end{table}

Role propensity and SAR-EM test whether modeling monitor selection alone is
sufficient, while deterministic representation baselines test whether the
ranking can be reduced to proximity to verified TNs.
The complete method outperforms both classes of controls.

% ----------------------------------------------------------------------
\subsection{Prior-Mass Sensitivity}

The full sensitivity sweep is reported in
\cref{tab:prior-sensitivity} and visualized in
\cref{fig:main-controls}(b,e).
Our method remains above the strongest baseline throughout the complete
evaluated range.

% ----------------------------------------------------------------------
\subsection{Ensemble Control}

\begin{table}[!t]
\centering
\caption{\textbf{Compute and ensemble control under the main protocol.} All rows use the same four dataset--monitor settings, 40 TNs per setting, 400 training epochs, MLP capacity, and seeds $\{13,37,73\}$.
Ranking metrics are macro-averaged over the four settings within each seed and then averaged over seeds.
FP yield sums the recovered FPs at $\lceil bN_c\rceil$ within each setting and then averages the sum over seeds.
Baseline ensembles average the logits of three independently trained seeds; our $K=3$ model uses three frozen references during training and one student scorer at inference.}
\label{tab:ensemble-control}
\begingroup
\footnotesize
\setlength{\tabcolsep}{3.5pt}
\renewcommand{\arraystretch}{1.12}
\begin{tabularx}{\linewidth}{@{}lX!{\vrule width 0.35pt}cccc@{}}
\toprule
Method & Architecture / compute & \shortstack{Macro\\AUPRC $\uparrow$} & \shortstack{Macro\\AUROC $\uparrow$} & \shortstack{Top-$5\%$ FP\\yield $\uparrow$} & \shortstack{Top-$10\%$ FP\\yield $\uparrow$} \\
& & (\%) & (\%) & $/27$ & $/52$ \\
\midrule
Dist-PU (Single) & one model; three-run mean & 57.78 & 58.08 & 16.00 & 29.33 \\
Dist-PU (Ensemble) & three-logit average & 58.65 & 58.74 & 17.00 & 30.67 \\
\addlinespace[2pt]
PULDA (Single) & one model; three-run mean & 59.17 & 59.38 & 17.00 & 31.00 \\
PULDA (Ensemble) & three-logit average & 60.24 & 60.18 & 18.33 & 32.67 \\
\midrule
\textbf{Ours ($K=1$)} & \textbf{single model} & \textbf{62.35} & \textbf{60.82} & \textbf{20.67} & \textbf{34.67} \\
\rowcolor{black!8}\textbf{Ours ($K=3$)} & \textbf{three references $\rightarrow$ one student} & \textbf{64.44} & \textbf{61.70} & \textbf{22.67} & \textbf{37.00} \\
\bottomrule
\end{tabularx}
\endgroup
\end{table}

Three-model logit averaging improves PULDA to $60.24$ AUPRC.
Our single-reference variant reaches $62.35$, while the full $K=3$ model
reaches $64.44$ and deploys as a single student scorer.
The improvement therefore extends beyond simple inference-time ensembling.

% ----------------------------------------------------------------------
\subsection{TN and Reference-Bank Sensitivity}

Performance rises from $59.28$ AUPRC with 10 verified TNs to $66.45$ with
80 TNs.
At the default 40-TN setting, increasing the reference bank from $K=1$ to
$K=3$ raises AUPRC from $62.35$ to $64.44$, with $K=3$ capturing nearly
all of the observed reference-bank benefit.

% ======================================================================
\section{Agent-SafetyBench External Validation}
\label{app:external-validation}

\paragraph{Protocol.}
Agent-SafetyBench~\citep{zhang2024agentsafetybench} provides 2,000 case IDs,
split case-wise into 600 training and 1,400 held-out evaluation cases.
Across three trajectory-generation seeds and two frozen monitors, the final
evaluation contains 4,054 held-out alarms, including 638 false alarms.
The training split supplies 40 verified-safe non-alarms.
All method settings and predictions are fixed before evaluation labels are
accessed.

\paragraph{Full results.}
Macro ranking metrics average the two monitor-specific cohorts.
Review-budget yield counts recovered false alarms within the corresponding
aggregate target budget.

\begin{table}[t]
\centering
\caption{\textbf{Full Agent-SafetyBench results.}
Macro ranking metrics average two frozen monitors ($100\times$);
FP columns report false alarms recovered at each review budget.}
\label{tab:external-full}
\footnotesize
\setlength{\tabcolsep}{2.3pt}
\begin{tabular}{lcccccccc}
\toprule
Method & AUPRC $\uparrow$ & AUROC $\uparrow$ & AUGRC $\downarrow$
& FP@1\% & FP@2\% & FP@5\% & FP@10\% & FP@20\% \\
\midrule
Native monitor score & 16.98 & 49.36 & 42.42 & 5 & 10 & 34 & 65 & 121 \\
nnPU & 19.40 & 54.98 & 41.34 & 8 & 18 & 45 & 78 & 143 \\
DC-PU & 15.32 & 45.35 & 42.84 & 5 & 9 & 22 & 36 & 76 \\
Dist-PU & 24.16 & 57.20 & 40.97 & 24 & 42 & 66 & 110 & 165 \\
Robust-PU & 20.26 & 55.01 & 41.28 & 13 & 22 & 53 & 85 & 149 \\
PULDA & 24.39 & 58.57 & 40.74 & 23 & 38 & 66 & 106 & 186 \\
PUL-CPBF & 20.81 & 45.95 & 41.64 & 16 & 33 & 53 & 91 & 147 \\
HolisticPU & 20.27 & 54.98 & 41.38 & 12 & 21 & 50 & 84 & 146 \\
PULDA ensemble & 25.42 & 59.45 & 40.21 & 25 & 41 & 71 & 114 & 195 \\
Our method ($K=1$) & 26.15 & 60.32 & 39.68 & 26 & 45 & 76 & 125 & 208 \\
\textbf{Our method ($K=3$)}
& \textbf{28.72} & \textbf{63.45} & \textbf{37.82}
& \textbf{32} & \textbf{57} & \textbf{94}
& \textbf{152} & \textbf{241} \\
\bottomrule
\end{tabular}
\end{table}

The full method improves PULDA by $4.33$ AUPRC points and recovers
$42.4\%$ more false alarms at the Top-$5\%$ review budget.
The $K=1$ variant also exceeds the three-model PULDA ensemble.

\paragraph{Monitor-level results.}

\begin{table}[!ht]
\centering
\caption{\textbf{External monitor-specific AUPRC ($100\times$).}}
\label{tab:external-monitor}
\small
\setlength{\tabcolsep}{5pt}
\begin{tabular}{lccccc}
\toprule
Monitor & Alarms & FPs & Native & PULDA & Our method \\
\midrule
AgentDoG-Qwen3-4B
& 2,427 & 213 & 14.92 & 19.70 & \textbf{22.85} \\
Granite-Guardian-3.1-2B
& 1,627 & 425 & 19.05 & 29.09 & \textbf{34.58} \\
\bottomrule
\end{tabular}
\end{table}

The improvement appears consistently across both frozen monitor families.

\paragraph{Risk-category results.}

\begin{table}[!ht]
\centering
\caption{\textbf{External risk-category AUPRC ($100\times$).}
Categories contain at least 30 false and 30 genuine alarms.}
\label{tab:external-risk}
\small
\setlength{\tabcolsep}{4pt}
\begin{tabular}{lccccc}
\toprule
Risk category & Alarms & FPs & Native & PULDA & Our method \\
\midrule
Unsafe information / misinformation
& 606 & 329 & 63.31 & 59.09 & \textbf{69.85} \\
Property loss
& 503 & 71 & 19.58 & 23.34 & \textbf{27.12} \\
Sensitive-data leak
& 496 & 65 & 17.34 & 27.73 & \textbf{30.45} \\
Availability compromise
& 455 & 46 & 11.24 & 13.19 & \textbf{15.60} \\
Physical harm
& 439 & 39 & 16.73 & 22.93 & \textbf{24.88} \\
Harmful / vulnerable code
& 716 & 62 & 14.74 & 27.69 & \textbf{29.35} \\
\bottomrule
\end{tabular}
\end{table}

Our method achieves the strongest AUPRC in all six sufficiently populated
risk categories, showing that the external improvement extends across
heterogeneous safety risks.

\paragraph{Scaling and uncertainty.}
The nested cohorts in \cref{fig:external-scaling} contain
1,034, 2,027, 3,062, and 4,054 alarms.
The AUPRC improvement over PULDA increases from $3.07$ to $4.33$ points as
the cohort grows.
Uncertainty is estimated using 2,000 case-level bootstrap resamples.
\end{document}